# MRS-YOLO: Railroad Transmission Line Foreign Object Detection Based on Improved YOLO11 and Channel Pruning

Siyuan Liu,[a] Junting Lin,[a,b,*]

[a]Lanzhou Jiaotong University, School of Automation and Electrical Engineering, Lanzhou, China, 730070
[b]Department of Mathematics and Statistics, University of Strathclyde, Glasgow G1 1XH, UK

**Abstract**. Aiming at the problems of missed detection, false detection and low detection efficiency in transmission line foreign object detection under railway environment, we proposed an improved algorithm MRS-YOLO based on YOLO11. Firstly, a multi-scale Adaptive Kernel Depth Feature Fusion (MAKDF) module is proposed and fused with the C3k2 module to form C3k2_MAKDF, which enhances the model's feature extraction capability for foreign objects of different sizes and shapes. Secondly, a novel Re-calibration Feature Fusion Pyramid Network (RCFPN) is designed as a neck structure to enhance the model's ability to integrate and utilize multi-level features effectively. Then, Spatial and Channel Reconstruction Detect Head (SC_Detect) based on spatial and channel preprocessing is designed to enhance the model's overall detection performance. Finally, the channel pruning technique is used to reduce the redundancy of the improved model, drastically reduce Parameters and Giga Floating Point Operations Per Second (GFLOPs), and improve the detection efficiency. The experimental results show that the mAP50 and mAP50:95 of the MRS-YOLO algorithm proposed in this paper are improved to 94.8% and 86.4%, respectively, which are 0.7 and 2.3 percentage points higher compared to the baseline, while Parameters and GFLOPs are reduced by 44.2% and 17.5%, respectively. It is demonstrated that the improved algorithm can be better applied to the task of foreign object detection in railroad transmission lines.

*Address all correspondence to Junting Lin**, E-mail: linjt@lzjtu.edu.cn

## 1. Introduction

With the advancement of high-speed railroad network intelligence, railroad transmission line as the core carrier of traction power supply system, its operation status directly affects the train scheduling safety and regional power supply stability, is the lifeblood of modern railroad transportation system[1]. In the complex and changing railroad operating environment, lightweight foreign objects such as plastic bags[2], floating kites, fabric materials[3] and balloons[4] are susceptible to entanglement on the line due to air currents, and the nesting behavior of birds[5] also creates a safety hazard. The intrusion of these foreign objects may not only trigger a short circuit tripping of the contact network, but also cause extensive train delays.

In the supervision of railroad transmission lines, conventional manual examination suffers from long time-consuming and high cost, which can't complete the task quickly and effectively in the face of the huge workload. Advancements in computational power and deep neural networks have enabled automatic detection technologies based on devices such as drones and vehicle-mounted video surveillance have received increasing attention due to their high efficiency and low cost[6-10]. Initially, deep learning based object detection was some two-stage algorithms like R-CNN series[11-13]. Later, the single-stage detection YOLO series algorithms[14-20] discarded the step of candidate region generation, solved the problems of slow detection speed and large computation, and were widely used in real-time detection tasks.

More and more scholars have improved the YOLO series of algorithms and applied them to different fields. For example, Wu et al.[21] proposed a lightweight remote sensing image object detection algorithm CBGS-YOLO based on YOLOv5, which improves the detection performance of small targets and reduces Parameters. Yu et al [22] improved YOLOv7 to enhance the performance of the algorithm for detecting foreign objects on transmission lines. Wang et al.[23] proposed an improved algorithm E-YOLO based on YOLOv8, which can efficiently detect estrus cow. Wang et al.[24] proposed an improved algorithm AG-YOLO based on YOLOv10.

In the face of the complex environment of the railroad, as well as the different sizes and shapes of foreign objects, the traditional object detection algorithm will be interfered with, and the phenomenon of misdetection and omission will occur. Currently, some scholars have conducted research on the issue of insufficient performance in detecting foreign objects on railway power transmission lines. For example, Hao et al.[25] proposed an improved algorithm called YOLO-LAF based on YOLOv8. Chen et al.[26] proposed the EPRepSADet detection algorithm. Their improvements have enhanced the accuracy of the algorithms in detecting foreign objects on power transmission lines in railway scenarios, but they are not lightweight enough. Since the detection algorithm needs to be deployed on edge devices such as drones and vehicle-mounted cameras for foreign object detection on power transmission lines, it is necessary to reduce parameters and GFLOPs while ensuring detection accuracy, thereby making the algorithm more lightweight. To address the above issues, this paper proposes an improved railway power line foreign object detection algorithm based on YOLO11, named MRS-YOLO. This algorithm not only improves the model's accuracy and reduces false positives and false negatives but also significantly reduces the model's complexity while maintaining accuracy, making the model more lightweight. The main contributions are as follows:

- Adaptive kernel depth convolution (AKDC) is proposed, and the proposed AKDC is utilized to construct a multi-scale adaptive kernel depth feature fusion module MAKDF, and MAKDF is incorporated into C3k2. Through the design of channel grouping and anisotropic convolutional kernel and the mechanism of generating channel-adaptive convolutional kernel weights, the network's ability to extract multi-scale features and focus on key regions is improved, which effectively enhances the model's ability to detect targets in challenging scenarios such as complex background and illumination transformation.
- A novel Recalibration Feature Fusion Pyramid Network (RCFPN) is designed to improve the neck network, which improves the feature fusion capability of the model through a more effective feature fusion method and enhances the boundary features of the target, which is conducive to the subsequent detection and classification of the object.
- A novel spatial and channel preprocessing based detect head SC_Detect is designed to improve the overall detection accuracy of the model by first preprocessing the target features input to the detector head with spatial and channel reconstruction, and then calculating the losses through the loss function.
- The improved model is lightened using the channel pruning technique to solve the problems of more redundant channels, excessive model computation, low detection speed, and large amount of Parameters when the improved model is performing target detection.

## 2. Method

### 2.1. YOLO11 algorithm

YOLO11 is an object detection algorithm proposed by Ultralytics in September 2024 based on YOLOv8. The YOLO series of algorithms treats object detection as a regression problem and is capable of simultaneous object localization and classification in a single image scan, combining high speed and accuracy. YOLO11 has five model sizes, from small to large, namely YOLO11n, YOLO11s, YOLO11m, YOLO11l, and YOLO11x. Its network structure consists of three parts: Backbone, Neck, and Head.

Backbone part for feature extraction, in which the C3K2 module extracts key information through an efficient cross-layer information fusion mechanism, the SPPF module can improve the multi-scale representation ability, and the C2PSA module improves the model's focus mechanism for salient areas by combining the spatial attention mechanism. The Neck part performs feature fusion, which employs a bottom-up feature pyramid network (FPN) for feature fusion to improve the detection performance. Head for detection output, YOLO11 compared to the previous version of the original convolution in the detection head is replaced by Depthwise Separable Convolution (DSConv), which reduces GFLOPs and parameters in the model, making the model more lightweight.

*2.2. MRS-YOLO algorithm*

In this paper, taking YOLO11n as the baseline, an improved algorithm MRS-YOLO is proposed for its shortcomings in the detection of foreign objects in railroad transmission lines, as shown in Fig. 1. Firstly, inspired by the ideas of GoogLeNet[27] and InceptionNeXt[28], this paper proposes the AKDC module, and designs the MAKDF module based on the AKDC module. The MAKDF module is integrated into C3K2 to form C3K2_MAKDF, replacing part of C3K2 in Backbone, which enhances the model's ability to extract features of foreign objects of different sizes and shapes. Secondly, the SBA module[29] is introduced into Neck, and a new FPN structure is proposed. The SBA module and C3K2_MAKDF are combined to design the RCFPN structure, which enhances the model's feature fusion ability. Then, ScConv[30] is introduced into the detection head to preprocess the input features and improve the detection accuracy. The fused detection head is named Sc_Detect. Finally, to address the computational inefficiencies caused by excessive GFLOP and Parameters of the improved model, the channel pruning technology is used to lightweight the improved model, which greatly reduces the amount of GFLOPs and Parameters at the expense of a small amount of accuracy. These improvements enable the MRS-YOLO algorithm to perform well in the task of detection in the face of images of railroad transmission lines containing foreign objects.

*2.2.1 C3K2_MAKDF module*

Szegedy et al. proposed the GoogLeNet[27] model

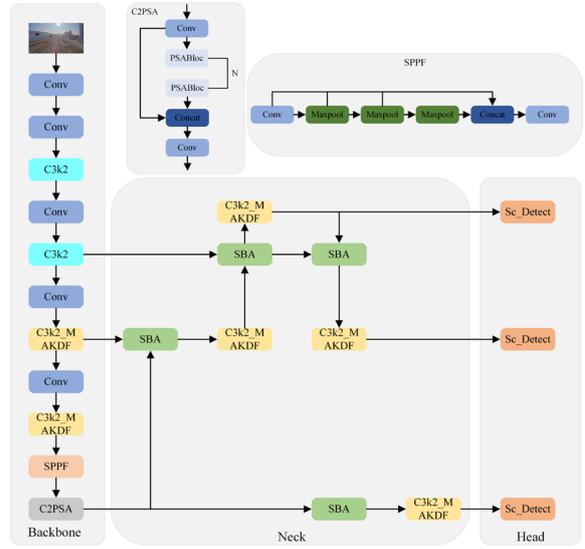

**Fig. 1** MRS-YOLO algorithm overall network structure diagram.

in 2014 a key part of which is the multi-scale grouped convolutional Inception architecture, which both improves the performance of the model and reduces the Parameters. The design of Inception is inspired by the idea of multi-scale feature fusion and sparse connection densification, where the input features are divided into four parallel paths by channel, namely 1x1 convolution, 3x3 convolution, 5x5 convolution and 3x3 maximum pooling, and the outputs of each branch are spliced together in the channel dimensions to form a multi-scale fusion of the feature maps, which significantly enhance the model's representational capacity.

Subsequently, Yu et al. proposed the InceptionNeXt module[28] on the idea of Inception. InceptionNeXt, in order to solve the bottleneck of the efficiency of large-core convolution, decomposes the large core into multiple groups of small cores branching in parallel, and introduces channel grouping strategy, which accelerates the large-core convolution without sacrificing the performance, and realizes a performance-computation-efficiency Balance between performance and computational efficiency.

Inspired by the ideas of Inception module and InceptionNeXt module, and considering that direct parallel convolution of input features grouped by channel may lose some important information in the channel, this paper proposes adaptive kernel deep convolution AKDC. The multi-branching mechanism is first set up to capture spatial features of different orientations and scales using square convolution ( $K \times K$ ), horizontal banding convolution ( $1 \times M$ ), and vertical banding convolution ( $M \times 1$ ) simultaneously with the input features, where $M = 3K + 2$. Then the weight coefficients of each branch are generated by adaptive average pooling, and the outputs of each branch are weighted and fused after Softmax normalization to achieve adaptive weighted fusion and retain important information, and the AKDC is shown in Fig. 2. Using the proposed AKDC combined with the parallel grouping idea, the multi-scale adaptive kernel depth feature fusion module MAKDF is constructed and shown in Fig. 3.

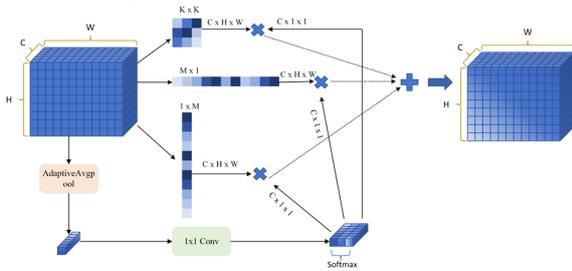

**Fig. 2** AKDC structure diagram.

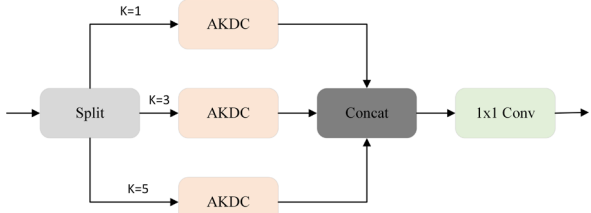

**Fig. 3** MAKDF structure diagram.

The multiscale adaptive kernel depth feature fusion module first divides the input features into three groups by channel, which is expressed by the formula:

$$(I_1, I_2, I_3) = Split(I, 3) \qquad (1)$$

where $I$ represents the input features, $Split(\cdot)$ represents the channel segmentation operation, and $I_1, I_2, I_3$ represents the three sets of segmented channels.

Then, the segmented three groups are passed into AKDC for feature extraction according to the rules of K=1, K=3 and K=5 respectively, enabling the model to learn features of different sizes, which is expressed by the formula:

$$\begin{cases} F_1 = A_1(I_1) \\ F_2 = A_3(I_2) \\ F_3 = A_5(I_3) \end{cases} \qquad (2)$$

where $F$ denotes the feature that has been processed by AKDC and $= A_n$ denotes $K = n$ in AKDC.

Finally, the features of each branch that have been processed by AKDC are aggregated by Concat, and then the features are fused to the output by a 1x1 point-by-point convolution, which is denoted by the formula:

$$O = C_{1x1}\left(Concat(F_1, F_2, F_3)\right) \qquad (3)$$

where $O$ denotes the output information after feature fusion, $C_{1x1}$ denotes the 1x1 convolution and $Concat(\cdot)$ denotes the aggregation operation.

The designed MAKDF module is incorporated into the C3k2 module in the original model. Replace the last convolution in the Bottleneck structure with the MAKDF module, and the improved module is named C3K2_MAKDF, as shown in Fig. 4.

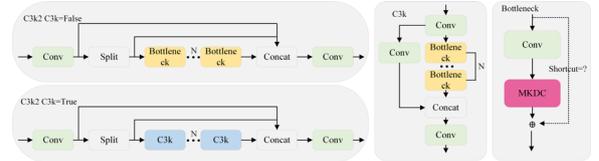

**Fig. 4** C3k2_MAKDF structure diagram.

The above-designed C3k2_MAKDF component empowers the network to more effectively extract feature information across varying sizes and morphologies, and adaptively retains more effective information, which enables the model to better accomplish the detection task in complex situations and enhances the model's generalization ability.

*2.2.2 RCFPN*

When detecting foreign objects in a railroad scene, due to the complex background environment and multiple downsampling, the features of the target foreign object may no longer be obvious enough or even blend into the background. Feature maps are less semantic in shallow networks, but rich in details and have more obvious boundaries; while deep networks contain rich semantic information, and the fusion method of direct splicing or summing is prone to information redundancy and feature conflicts. In order to better integrate low-level features and high-level features, the Selective Boundary Aggregation (SBA) module proposed by the Dual-Aggregation Transformer (DuAT)[29] is introduced. The RCFPN is designed in combination with the designed C3K2_MAKDF module to improve the neck network. The SBA module aims to solve the problems of easy loss of boundary details and redundancy in cross-level feature fusion in target detection tasks. It proposes a mechanism based on cross-level feature complementary fusion and adaptive selective calibration. By combining the detail expression ability of low-level feature representations and high-level semantic abstractions, it achieves the coordinated optimization of boundary accuracy and semantic integrity.

The key part of the SBA module is Re-calibration Attention Unit (RAU). The SBA module uses two RAU units to perform bidirectional guided calibration of shallow and deep features, embeds deep semantic priors in the shallow path to enhance target discrimination ability, and injects shallow boundary constraints in the deep path to repair contour distortion. RAU adopts a dual-branch attention collaborative strategy to solve the problem of feature misalignment. Firstly, the multi-head attention mechanism is utilized to compute the cross-resolution feature correlation, and then the feature response weights are dynamically assigned by the gating function to strengthen the boundary response in the shallow branch, and amplify the semantically significant region in the deep branch in the middle.

The SBA module first adaptively extracts complementary representations from the two input features $F_H$ and $F_L$. The shallow and deep features are connected to the RAU through differentiation. Then, the RAU-processed deep features are upsampled to match the size of the target feature map. Finally, the outputs of the two branches are spliced before performing a 3x3 convolutional output. The function $R(\cdot)$ of the RAU module can be expressed as:

$$F_1' = S_\theta(F_1), F_2' = S_\delta(F_2) \quad (4)$$

$$R(F_1, F_2) = F_1' \odot F_1 + F_2' \odot F_2 \odot \left(\ominus\left(F_1'\right)\right) + F_1 \quad (5)$$

where $F_1$ and $F_2$ represent two input features. $S_\theta$ and $S_\delta$ represent two linear mappings and Sigmoid functions, which are used to obtain feature maps $F_1'$ and $F_2'$. $\odot$ represents the dot product. $\ominus\left(F_1'\right)$ represents the inverse operation of subtracting $F_1'$.

The whole process of the SBA module can be represented as follows:

$$A = C_{3\times 3}\left(Concat\left(R(F_H, F_L), R(F_L, F_H)\right)\right) \quad (6)$$

where $A$ denotes the output of the SBA module.

The RCFPN structure is designed by combining the SBA module with the C3K2_MAKDF module, as shown in Fig. 5. The design allows Neck to autonomously suppress redundant background interference and enhance the boundary features of the target when performing feature fusion, while retaining the complementary benefits of multi-scale information.

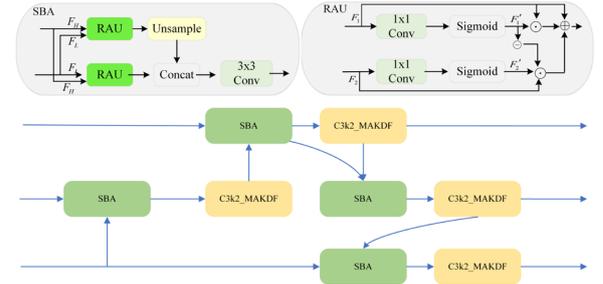

**Fig. 5** RCFPN structure diagram.

*2.2.3 Sc_Detect*

In object detection, the features obtained from the image to be detected after feature extraction and feature fusion will be transmitted to the head. The performance of the head will directly affect the final detection accuracy. To strengthen the adaptive ability of the detection head to deal with different kinds of foreign objects and improve the detection effect, we combine the original detection head of YOLO11 with the spatial and channel reconstruction convolution ScConv[30] proposed by Li et al. and proposes Sc_Detect, as shown in Fig. 6.

The features input to the detection head are first pre-processed spatially and channel-wise by the ScConv module, and then convolved and loss is calculated. By reconstructing the spatial and channel information of the feature map, the correlation between different locations and different channels is better captured, thus eliminating feature redundancies while enhancing discriminative power.

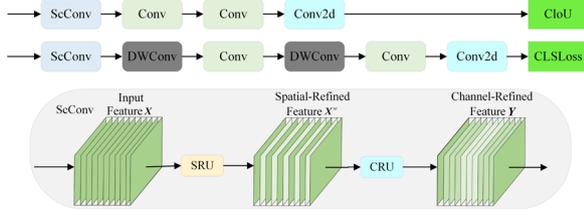

**Fig. 6** Sc_Detect structure diagram.

ScConv is mainly composed of spatial reconstruction unit (SRU) and channel reconstruction unit (CRU). The input features are initially processed by the SRU module to generate spatially optimized representations, which are subsequently fed into the CRU module for yielding channel-optimized features.

The separation process of SRU first evaluates the information content in different feature maps using Group Normalization (GN) to discriminate between high-utility and low-utility features, and then maps to (0, 1.0) by Sigmoid and sets a threshold for gating. When the input feature is $X$, the transformation is formally defined as:

$$X_{out} = \text{GN}(X) = \lambda \frac{X - \mu}{\sqrt{\sigma^2 + \varepsilon}} + \eta \quad (7)$$

$$w_i = \lambda_i / \sum_{j=1}^{H} \lambda_j; i, j = 1, 2 \cdots H \quad (8)$$

$$W = \text{Gate}\left(Sigmoid\left(W_\lambda\left(\text{GN}(X)\right)\right)\right) \quad (9)$$

where $\mu$ and $\sigma$ denote the mean and standard deviation. $\lambda$ and $\eta$ denote trainable variables. $\varepsilon$ is a minimal constant that ensures stability. $w_i$ is an element in $W_\lambda$. $H$ is the number of channels.

After separation two weighted features $X_1^w$ and $X_2^w$ are obtained, where $X_1^w$ has more spatial content and $X_2^w$ has less information. $X_1^w$ and $X_2^w$ are further divided into two parts each $X_{11}^w$, $X_{12}^w$ and $X_{21}^w$, $X_{22}^w$. The spatial refinement feature $X^w$ is then obtained by using cross-reconstruction and splicing, which is denoted by the formula:

$$\begin{cases} X_1^w = W_1 \odot X, \quad X_2^w = W_2 \odot X \\ X_{11}^w \oplus X_{22}^w = X^{w1}, \quad X_{21}^w \oplus X_{12}^w = X^{w2} \\ X^{w1} \cup X^{w2} = X^w \end{cases} \quad (10)$$

where $\oplus$ denotes the element addition and $\cup$ denotes the matrix splice operation.

CRU is mainly used for channel-wise redundancy reduction, improve computational efficiency, and enhance representative features. First, the $X^w$ generated by SRU is divided into two parts, and $X_{up}$ and $X_{low}$ are obtained by convolution. The two features are then combined by Global Weighted Convolution (GWC) and Partial Weighted Convolution (PWC) processing into $Y_1$ and $Y_2$. The formula is expressed as:

$$Y_1 = M^G X_{up} + M^P X_{up} \quad (11)$$

$$Y_2 = M^P X_{low} \cup X_{low} \quad (12)$$

where $M^G$ and $M^P$ denote the learnable matrices of GWC and PWC, respectively.

Finally, a simplified SKNet method is used for adaptive fusion of $Y_1$ and $Y_2$. $S_1$ and $S_2$ are obtained using average pooling, followed by Softmax operation to obtain feature weight vectors $\eta_1$ and $\eta_2$, and then fused according to the weights to obtain the final feature $Y$. The formula is denoted as:

$$S_m = \text{Pooling}(Y_m) \quad (13)$$

$$\eta_1 = \frac{e^{s_1}}{e^{s_1} + e^{s_2}}, \eta_2 = \frac{e^{s_2}}{e^{s_1} + e^{s_2}}, \eta_1 + \eta_2 = 1 \quad (14)$$

$$Y = \eta_1 Y_1 + \eta_2 Y_2 \quad (15)$$

### 2.2.4 Channel pruning

Deep neural networks are currently showing superior performance in various domains, but they also have huge memory and computational power requirements. In order to deploy deep neural networks in limited hardware resources and maximize their advantages, pruning algorithms are gradually being widely used. Pruning is the process of removing redundant parts of the network model that are unimportant and take up a lot of resources, drastically reducing GFLOPs and Parameters with little or no impact on the accuracy.

The improved model improves the detection accuracy, but at the same time increases Parameters and GFLOPs of the model. In order to lighten the improved model, this paper uses the Layer-Adaptive Magnitude-based Pruning (LAMP) algorithm[31] to prune the model. In a neural network, each connection has a weight. With this weight, the LAMP algorithm can get a LAMP score, and then the model can be lightened by removing the parts with smaller scores. The calculation of LAMP score is denoted as:

$$\text{score}(u;W) = \frac{(W[u])^2}{\sum_{v \geq u}(W[v])^2} \quad (16)$$

where $\text{score}(\cdot)$ denotes the LAMP score. $W[v]$ denotes the weight of the target connection. $W[u]$ denotes the weight of all remaining connections in the same layer.

LAMP scores are calculated and ranked for all connections in each layer, and those with lower LAMP scores are considered unimportant parts and are removed.

## 3. Experimentation and Analysis

### 3.1. Experimental dataset

The experiments use RailFOD23[32], a publicly available dataset for foreign object detection on railroad transmission lines. It contains 14615 images with a total of 40541 labeled objects, which contain four common categories: balloons, floats, bird's nests, and plastic bags. The dataset is divided into training and validation sets in the ratio of 8:2.

### 3.2. Evaluation Metrics

In order to verify the performance of the model, this experiment adopts mean Average Precision (mAP) as an evaluation index in terms of detection accuracy, the higher the mAP the better the detection effect, which is calculated by the precision P and the recall rate R, which is calculated by the formula:

$$P = \frac{TP}{TP+FP} \quad (17)$$

$$R = \frac{TP}{TP+FN} \quad (18)$$

$$AP_i = \int_0^1 P_i(R)\,dR \quad (19)$$

$$mAP = \frac{\sum_{i=1}^{n} AP_i}{n} \quad (20)$$

where $AP_i$ denotes the average precision of the $i$-th category.

Parameters and GFLOPs serve as quantitative indicators of model complexity. Parameters represents the spatial complexity, the smaller the Parameters, the easier the model is to be deployed. GFLOPs represents the temporal complexity, the smaller the GFLOPs, the faster the detection speed.

### 3.3. Experimental Environment

The system environment used for the experiment was Ubuntu 22.04, the central processing unit was Intel(R) Core(TM) i9-13900KF, the graphics processing unit was NVIDIA GeForce RTX 4090, the deep learning architecture was pytorch1.8.0+cu111. The model's parameter settings are shown in Table 1.

**Table 1** Model parameter setting.

| parameter name | setting |
|---|---|
| categories | 4 |
| learning rate | 0.01 |
| workers | 8 |
| epochs | 200 |
| batch | 32 |

### 3.4. Analysis of experimental results

#### 3.4.1 SC_Detect module experiment

The threshold weights of the SRU unit in the improved SC_Detect module need to be gated. Experiments were conducted by setting 9 thresholds (0.1-0.9) to determine the optimal parameters. As shown in Table 2. mAP50 denotes the average precision when the IoU threshold is 0.5, and mAP50:95 denotes the average precision in the range of IoU thresholds from 0.5 to 0.95.

Table 2 SC_Detect module ablation experiment.

| ratios | mAP50/% | mAP50:95/% | Parameters | GFLOPs |
|---|---|---|---|---|
| baseline | 94.1 | 84.1 | 2,582,932 | 6.3 |
| 0.1 | 94.2 | 84.7 | 2,735,739 | 6.5 |
| 0.2 | 94.3 | 84.9 | 2,735,739 | 6.5 |
| 0.3 | 94.3 | 85.1 | 2,735,739 | 6.5 |
| **0.4** | **94.4** | **85.2** | **2,735,739** | **6.5** |
| 0.5 | 94.2 | 84.7 | 2,735,739 | 6.5 |
| 0.6 | 94.2 | 84.9 | 2,735,739 | 6.5 |
| 0.7 | 94.2 | 84.4 | 2,735,739 | 6.5 |
| 0.8 | 94.2 | 84.5 | 2,735,739 | 6.5 |
| 0.9 | 94.2 | 84.4 | 2,735,739 | 6.5 |

Table 2 shows that the Parameters and GFLOPs are unchanged for the nine cases, with a small increase compared to Baseline. The accuracy is increased. Where the accuracy achieves the optimal value when the ratio is 0.4.

*3.4.2 Module ablation experiment*

In this paper, the MRS-YOLO model is constructed based on the YOLO11n with 3 improvements. As shown in Table 3. The ablation experiments were carried out on the RailFOD23 dataset to verify the effectiveness of each module. Where M denotes the replacement of the C3k2 module in the backbone network with C3K2_MAKDF, R means replacing the neck with RCFPN, and S denotes the replacement of the original detection header with SC_Detect.

The results in Table 3 show that the three modules can improve the accuracy of the model. Adding each module improves both mAP50 and mAP50:95. The Parameters and GFLOPs of the model also decrease when C3K2_MAKDF is added, but both Parameters and GFLOPs increase somewhat when the RCFPN module and SC_Detect module are added. Finally, adding three modules simultaneously, the MRS-YOLO model proposed in this study, improved the mAP50 and mAP50:95 by 0.9 and 2.6 percentage points, respectively.

*3.4.3 Channel pruning experiment*

As shown in Table 4. To address the challenge posed by high Parameters and GFLOPs of the improved model, the LAMP algorithm is used for pruning to remove the redundant parts of the model. To determine the optimal pruning rate, nine pruning rates ranging from 0.1 to 0.9 were experimentally set, and the trend is shown in Fig. 7.

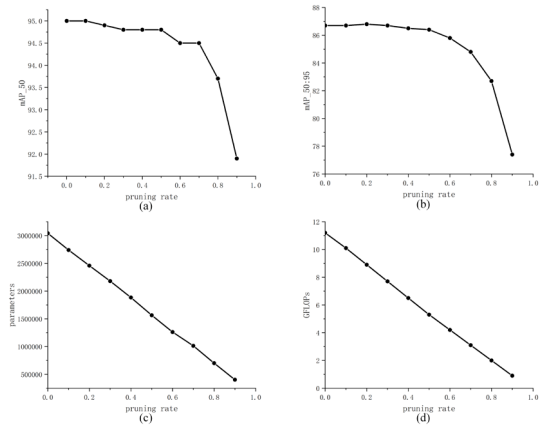

**Fig. 7** Trends in indicators at different pruning rates.

Table 4 Channel pruning experiment.

| pruning rates | mAP50/% | mAP50:95/% | Parameters | GFLOPs |
|---|---|---|---|---|
| 0 | 95 | 86.7 | 3,042,357 | 11.2 |
| 0.1 | 95 | 86.7 | 2,740,041 | 10.1 |
| 0.2 | 94.9 | 86.8 | 2,458,556 | 8.9 |
| 0.3 | 94.8 | 86.7 | 2,179,467 | 7.7 |
| 0.4 | 94.8 | 86.5 | 1,884,031 | 6.5 |
| **0.5** | **94.8** | **86.4** | **1,442,144** | **5.2** |
| 0.6 | 94.5 | 85.8 | 1,262,144 | 4.2 |
| 0.7 | 94.5 | 84.8 | 1,012,898 | 3.1 |
| 0.8 | 93.7 | 82.7 | 701,041 | 2.0 |
| 0.9 | 91.9 | 77.4 | 402,536 | 0.9 |

Table 3 Module ablation experiment.

| YOLO11n | M | R | S | mAP50/% | mAP50:95/% | Parameters | GFLOPs |
|---|---|---|---|---|---|---|---|
| √ | | | | 94.1 | 84.1 | 2,582,932 | 6.3 |
| √ | √ | | | 94.3 | 84.8 | 2,082,932 | 5.8 |
| √ | | √ | | 94.8 | 86.1 | 3,310,691 | 12.1 |
| √ | | | √ | 94.4 | 85.2 | 2,735,739 | 6.5 |
| √ | √ | √ | √ | 95 | 86.7 | 3,042,357 | 11.2 |

The trend graphs of the changes obtained by observing the experiments conducted by setting different pruning rates. With progressively higher pruning rates, Parameters and GFLOPs of the model decreases gradually. When the pruning rate is lower than 0.5, the change trend of mAP50 and mAP50:95 of the model is not significant, which has little effect on the effectiveness of detection. However, when the pruning rate is higher than 0.5, the mAP50 and mAP50:95 decrease faster, which has a greater impact on the effectiveness of detection. On the basis of ensuring the detection accuracy, the pruning rate is chosen to be 0.5, which can effectively reduce the Parameters and GFLOPs with little impact on the accuracy. Finally, after pruning, the mAP50 and mAP50:95 of the MRS-YOLO model proposed in this paper are 94.8% and 86.4%, respectively, and the Parameters and GFLOPs are 1,442,144 and 5.2, respectively. Comparing with the model before pruning, the accuracy is guaranteed, and it makes the Parameters and GFLOPs reduced by about 50%, respectively.

*3.4.4 Comparative experiment*

To verify the performance advantages of the proposed MRS-YOLO model in the task of detecting foreign objects in railroad transmission lines, comparison experiments are conducted between MRS-YOLO and several other models including the latest YOLO13 on the RailFOD23 dataset, and the results of the comparison experiments of different models are shown in Table 5.

MRS-YOLO improved mAP50 by 0.7% and mAP50:95 by 2.3% compared to the baseline YOLO11n model, Parameters decreased from 2,582,932 to 1,442,144, and GFLOPs decreased from 6.3 to 5.2. Compared to the latest YOLO12n model, mAP50 is improved by 0.6%, mAP50:95 is improved by 2.6%, and Parameters and GFLOPs are drastically reduced. The comparative test results further validate the advantages of the improved algorithm MRS-YOLO in terms of foreign object detection performance in railroad transmission line scenarios, which improves the detection accuracy while drastically reduces the Parameters and GFLOPs of the model to improve the detection efficiency.

**Table 5** Comparative experiments with different algorithms.

| algorithms | mAP50/% | mAP50:95/% | Parameters | GFLOPs |
|---|---|---|---|---|
| YOLOv6[14] | 94.1 | 83.7 | 4,234,635 | 11.8 |
| YOLOv7-tiny[15] | 94.0 | 79.5 | 6,036,636 | 13.2 |
| YOLOv8n[16] | 93.9 | 83.8 | 3,006,428 | 8.1 |
| YOLOv10n[18] | 93.8 | 82.8 | 2,266,923 | 6.5 |
| YOLO11n[19] | 94.1 | 84.1 | 2,582,932 | 6.3 |
| YOLO12n[20] | 94.2 | 83.8 | 2,557,508 | 6.3 |
| YOLO13n | 94.1 | 83.9 | 2,449,650 | 6.2 |
| **MRS-YOLO** | **94.8** | **86.4** | **1,442,144** | **5.2** |

*3.4.5 Visualization of results*

For visual comparison of the detection effects of the MRS-YOLO model proposed in this paper and the baseline YOLO11n model, some images from the validation set are selected for validation, as shown in Fig. 8. As can be seen from the figure, MRS-YOLO's ability to detect foreign objects on the railroad transmission line is significantly better than that of the baseline YOLO11n. The first column of YOLO11n mistakenly detected the leaf in the upper left corner as a floating object, and MRS-YOLO did not detect it incorrectly. The second column YOLO11n did not detect the floating object on the transmission line and MRS-YOLO successfully detected it. The third and fourth columns of the YOLO11n were roughly detected in the face of a large number of cluttered foreign objects and had many missed detections, while the MRS-YOLO was more delicate and reduced the missed detection rate. In the fifth column, YOLO11n did not detect one of the green floating objects due to its small size and long shape, which was interfered by the background, and MRS-YOLO successfully detected it.

## 4. Conclusion

In this paper, an improved object detection algorithm MRS-YOLO based on YOLO11n is proposed to address the problems of missed and false detection of targets caused by the diverse shapes of images of foreign objects on railroad transmission lines, large scale variations, and strong background interference. First, the AKDC module is proposed, and the MAKDF module is designed based on the AKDC module, and the MAKDF module is integrated into C3K2, which enhances the feature extraction capability of the model for different sizes and shapes of foreign objects. Secondly, a novel RCFPN was designed by combining the SBA module and C3K2_MAKDF to enhance the feature fusion capability of the model. Then, ScConv is incorporated into the head, and the preprocessing detection head Sc_Detect is proposed to improve the detection accuracy. Finally, to address the problem of large Parameters and GFLOPs in the improved model, the improved model is lightened using channel pruning techniques to significantly reduce Parameters and GFLOPs at the expense of a small amount of accuracy. Compared with YOLO11n, the mAP50 of MRS-YOLO has increased from 94.1% to 94.8%, the mAP50:95 has increased from 84.1% to 86.4%, the Parameters have been reduced from 2582932 to 1564050, and the GFLOPs have been reduced from 6.3 to 5.2. The improved algorithm has improved both the detection accuracy and the detection efficiency, and has shown obvious advantages in the experimental results. In the future, the algorithm performance will be further optimized, and the model's foreign body detection capability in the railway transmission line scenario will be improved through technologies such as knowledge distillation.

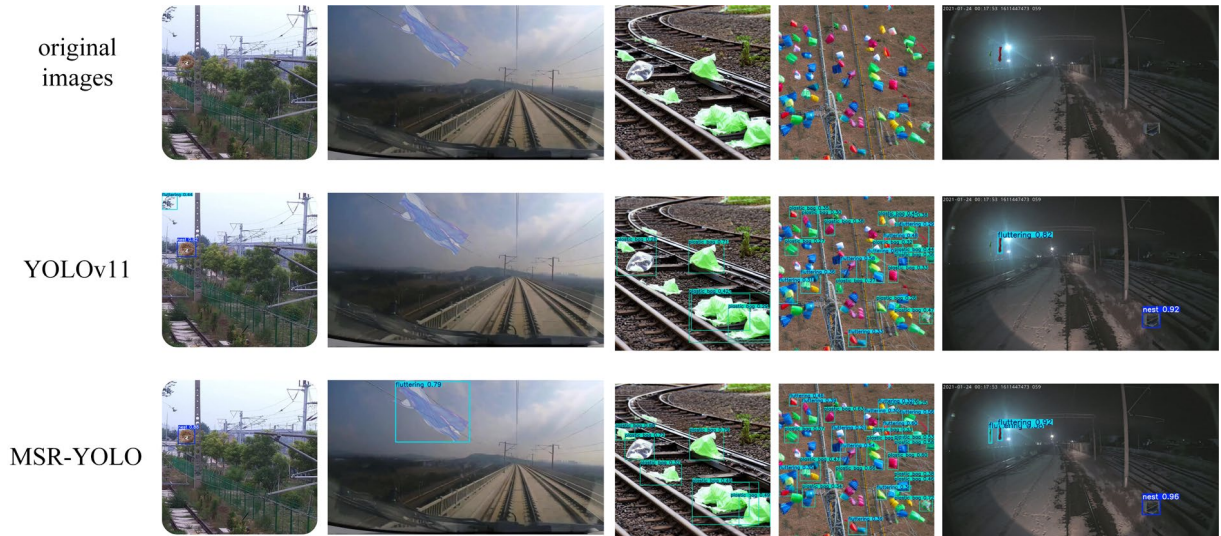

**Fig. 8** Visual comparison of YOLO11n and MRS-YOLO detection results.

## Acknowledgments

This work was supported by the National Natural Science Foundation of China under Grant 52162050 and the Center of National Railway Intelligent Transportation System Engineering and Technology (grant number RITS2025KF07).

## Data availability

The data supporting the findings of this study are publicly available datasets that can be accessed on the official website.